\documentclass[acmtog,screen]{acmart}

\usepackage{booktabs} 

\usepackage[utf8]{inputenc} 
\usepackage[T1]{fontenc}    
\usepackage{hyperref}       
\usepackage{url}            
\usepackage{amsfonts}       
\usepackage{nicefrac}       
\usepackage{microtype}      
\usepackage{xcolor}         
\usepackage{multirow}
\usepackage{textcomp}
\usepackage{wrapfig}
\usepackage{siunitx}

\usepackage[ruled]{algorithm2e} 

\SetAlFnt{\small}
\SetAlCapFnt{\small}
\SetAlCapNameFnt{\small}
\SetAlCapHSkip{0pt}

\acmJournal{TOG}
\copyrightyear{2023} 
\acmYear{2023} 
\setcopyright{rightsretained} 
\acmConference[SA Conference Papers '23]{SIGGRAPH Asia 2023 Conference Papers}{December 12--15, 2023}{Sydney, NSW, Australia}
\acmBooktitle{SIGGRAPH Asia 2023 Conference Papers (SA Conference Papers '23), December 12--15, 2023, Sydney, NSW, Australia}
\acmDOI{10.1145/3610548.3618228}
\acmISBN{979-8-4007-0315-7/23/12}

\newcommand{\projecturl}{https://samir55.github.io/3dshapematch/}
\newcommand{\projecthref}{\href{\projecturl}{\projecturl}}

\begin{document}
\title{Zero-Shot 3D Shape Correspondence}

\author{Ahmed Abdelreheem}
\affiliation{%
 \institution{KAUST}
 \country{Saudi Arabia}
 }
\email{ahmed.abdelreheem@kaust.edu.sa}
\author{Abdelrahman Eldesokey}
\affiliation{%
 \institution{KAUST}
 \country{Saudi Arabia}
}
\email{abdelrahman.eldesokey@kaust.edu.sa}

\author{Maks Ovsjanikov}
\affiliation{%
\institution{LIX, École Polytechnique}
\country{France}
}
\email{maks@lix.polytechnique.fr}

\author{Peter Wonka}
\affiliation{%
 \institution{KAUST}
 \country{Saudi Arabia}
}
\email{pwonka@gmail.com}

\begin{teaserfigure}
    \centering
    \includegraphics[width=\linewidth]{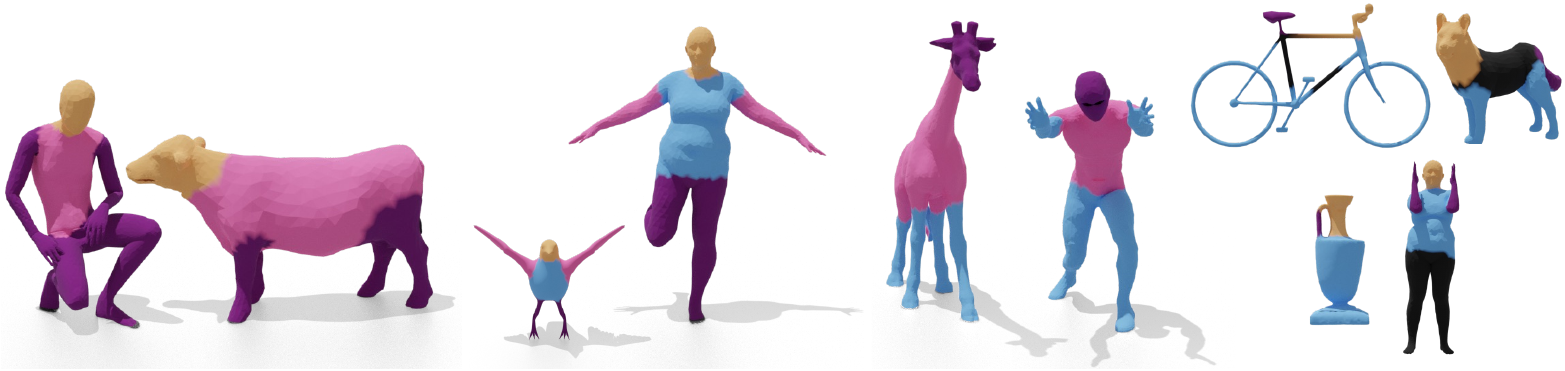}
    \caption[]{Our proposed approach can produce 3D shape correspondence maps for \emph{strongly non-isometric} shapes in a zero-shot manner. Mutual semantic regions are matched and are shown in similar colors, while non-mutual regions that can not be matched are shown in black. }
    \Description{This is the teaser figure of the proposed method. It contains pairs of shapes (for example, a dog and a bicycle shape pair) where similar semantic regions are matched (for example, matching the legs of the dog with the wheels of the bicycle).}
    \label{fig:teaser}
\end{teaserfigure}

\begin{abstract}

We propose a novel zero-shot approach to computing correspondences between 3D shapes.
Existing approaches mainly focus on isometric and near-isometric shape pairs (\textit{e}.\textit{g}., human vs. human), but less attention has been given to strongly \emph{non-isometric} and \emph{inter-class} shape matching  (\textit{e}.\textit{g}., human vs. cow).
To this end, we introduce a fully automatic method that exploits the exceptional reasoning capabilities of recent foundation models in language and vision to tackle difficult shape correspondence problems.
Our approach comprises multiple stages.
First, we classify the 3D shapes in a zero-shot manner by feeding rendered shape views to a language-vision model (\textit{e}.\textit{g}., BLIP2) to generate a list of class proposals per shape.
These proposals are unified into a single class per shape by employing the reasoning capabilities of ChatGPT.
Second, we attempt to segment the two shapes in a zero-shot manner, but in contrast to the co-segmentation problem, we do not require a mutual set of semantic regions.
Instead, we propose to exploit the in-context learning capabilities of ChatGPT to generate two different sets of \emph{semantic regions} for each shape and a \emph{semantic mapping} between them. 
This enables our approach to match strongly non-isometric shapes with significant differences in geometric structure.

Finally, we employ the generated semantic mapping to produce coarse correspondences that can further be refined by the functional maps framework to produce dense point-to-point maps.
Our approach\footnote{Project webpage: \projecthref}, despite its simplicity, produces highly plausible results in a zero-shot manner, especially between \emph{strongly non-isometric} shapes.

\end{abstract}

%
%

%
%


\maketitle

\section{Introduction}
Shape correspondence is a fundamental task in computer vision.
The objective of this task is to match two 3D shapes given some geometric representation (\textit{e}.\textit{g}., point clouds, meshes) to produce a region-level or point-level mapping.
This mapping can be constrained based on the downstream application in terms of deformation type, density, and scope (partial or full).
Examples of such downstream applications are shape interpolation, shape morphing, shape anomaly detection, 3D scan alignment, and motion capture. 
While early approaches for shape correspondence mainly adopted optimization-based algorithms \cite{kaick_survey_2010}, the emergence of deep learning has paved the way for learning-based approaches that implicitly learn suitable representations, which can be used for efficiently solving the matching problem.
These approaches can either follow a supervised \cite{litany_deep_2017}, unsupervised \cite{halimi_unsupervised_2019,cao2023unsupervised} or a self-supervised \cite{cao_self-supervised_2023} paradigm depending on the availability and the diversity of annotated datasets.

Supervised approaches are naturally data-dependent, and they require large-scale datasets with different classes of shapes for strong generalization.
On the other hand, unsupervised alternatives are class-agnostic and do not require annotated data, but they still lag behind their supervised counterparts in terms of performance \cite{halimi_unsupervised_2019}.
Moreover, the majority of these aforementioned approaches attempt to match near-isometric shapes of the same class, with less focus on non-isometric shapes (\textit{e}.\textit{g}., human v.s. animal). 
This is mainly caused by the lack of datasets with inter-class shape pairs and the complexity of matching dissimilar shapes.
To achieve a deeper understanding of 3D shapes and their relationships, it is desirable to develop methods that can generalize well both to isometric and non-isometric shape matching.

Recently, several large-scale models were introduced for different modalities such as language (\textit{e}.\textit{g}., GPT3 \cite{GPT3}, Bloom \cite{scao2022bloom}), and vision (\textit{e}.\textit{g}., StableDiffusion \cite{LDM}, DALLE-2 \cite{DALLE-2}).
These models are usually referred to as \emph{foundation models}, and they have a broad knowledge of their domains since they were trained on large amounts of data.
There are even ongoing efforts to connect these models to build a bridge between different modalities, such as Visual ChatGPT \cite{wu2023visual}, and MiniGPT-4 \cite{zhu2023minigpt}.
Unfortunately, it is still challenging to build similar models for modalities with limited amounts of data, such as 3D shapes.
Therefore, a plausible approach is to employ existing models for language and vision to solve problems for other data-limited modalities.

Motivated by this, we attempt to exploit existing foundation models to perform zero-shot shape correspondence with no additional training or finetuning. 
To address this problem, we identified the following three key problems.
First, we would like to predict the class of each of the two shapes in question given only their 3D meshes.
We achieve this through zero-shot shape classification by feeding rendered views of the two shapes into a language-visual model, BLIP2 \cite{li2023blip2}, to obtain object class proposals.
Then, we use ChatGPT to merge these proposals into a single class per object.
Second, we produce a semantic region set per shape, and a semantic mapping between the sets by exploiting the in-context learning \cite{GPT3} capabilities of ChatGPT \cite{gpt35}. 
Afterward, we introduce a zero-shot 3D semantic segmentation method based on the recent large-scale models DINO \cite{DINO} and Segment-Anything (SAM) \cite{kirillov2023segany}, which we denote as \emph{SAM-3D}. 
Our method only requires a shape mesh and its corresponding semantic region set as input. 
Finally, the semantic mapping is used to provide coarse correspondences between the two shapes and a finer map can be produced, if needed, by employing the functional map framework \cite{ovsjanikov_functional_2012}.
Remarkably, although functional maps are geared towards near-isometric shape pairs, we observe that it is possible to obtain high-quality dense maps given an initialization from SAM-3D, even across some challenging non-isometric shapes.

Since we propose a new scheme for solving the shape correspondence problem, we introduce several evaluation metrics to evaluate the performance of different intermediate tasks in our pipeline, such as zero-shot object classification, semantic region generation, and semantic segmentation. 
We also create a new benchmark that includes \emph{strongly non-isometric} shape pairs (\textit{e}.\textit{g}., humans vs. animals) that we denote as (SNIS) in order to test the generalization capabilities of our proposed approach.
Experiments on the new benchmark show that our approach, despite being zero-shot, performs very well on non-isometric  shape pairs.

\noindent In summary, we make the following contributions:
\begin{itemize}
    \item We propose a novel solution to 3D shape correspondence that computes results in a zero-shot manner.
    \item To the best of our knowledge, we introduce a first zero-shot joint 3D semantic segmentation technique that does not start with a mutual set of semantic regions, and it requires only the shape meshes while exploiting language-vision models to generate shape-specific semantic regions.    
    \item We introduce a benchmark for shape correspondence which includes strongly non-isometric shape pairs, as well as evaluation metrics for different stages of the proposed pipeline.
\end{itemize}


\section{Related Work}

In this section, we give a brief overview of shape correspondence literature, large-scale models, and 3D semantic segmentation. For shape correspondence, we focus only on relevant deep learning-based approaches, and we refer the reader to \cite{kaick_survey_2010,sahillioglu_recent_2020} for a comprehensive survey of earlier registration-based and similarity-based approaches.

\begin{figure*}[!htb]
    \centering
    \includegraphics[width=\linewidth]{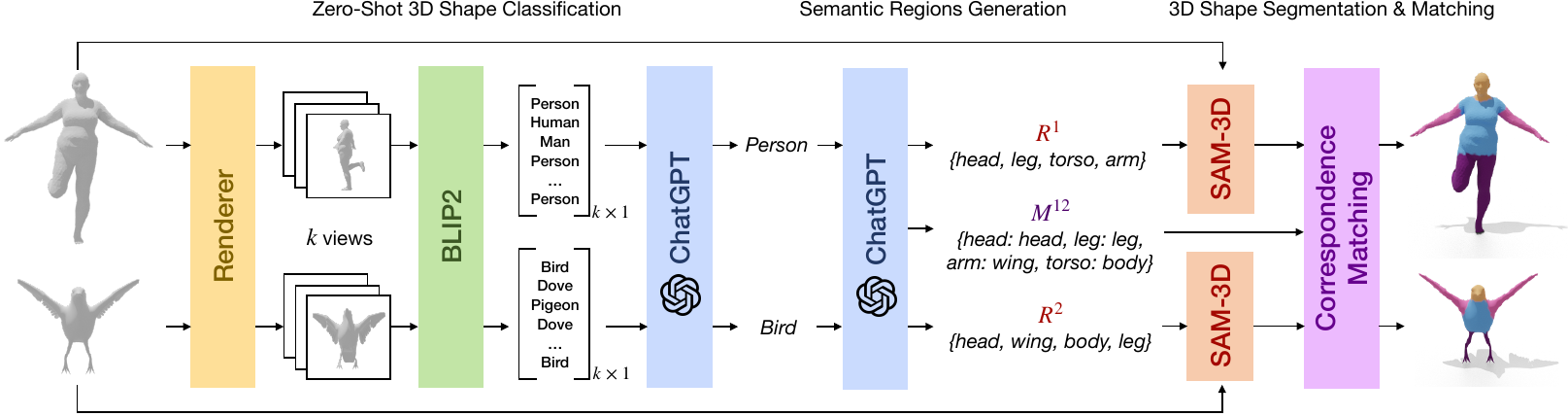}
    \caption[]{Our proposed approach has three main components: \textbf{(1) Zero-shot 3D shape classification}: By feeding rendered \textit{k} views of each shape to a BLIP2 \cite{li2023blip2} model to generate class proposal lists. The proposals are unified using ChatGPT to produce a single class per shape. \textbf{(2) Semantic region/mapping generation}: In-context learning capabilities of ChatGPT are employed to produce a semantic region set for each shape and a semantic mapping between them. \textbf{(3) Zero-shot 3D semantic segmentation}: our proposed SAM-3D uses the semantic regions to segment the shapes, and the mapping is used to produce a sparse correspondence map that can be densified further using the functional maps framework \cite{ovsjanikov_functional_2012}. }
    \Description{This figure shows the pipeline of the proposed method.}
    \label{fig:method12}
\end{figure*}

\subsection{Deep Learning-Based Shape Correspondence}
Convolutional Neural Networks (CNNs) by nature are not directly applicable to non-rigid shapes due to the lack of shift-invariance property in non-Euclidean domains.
Wei \textit{et al. } \cite{wei_dense_2016} circumvented this by training on depth maps of shapes that are being matched, and produced pixel-wise classification maps for each point in the object.
Wu \textit{et al. } \cite{wu_3d_2015} generated volumetric representations from depth maps, and used 3D CNNs to process them.
However, these methods do not capture all shape deformations, since they treat shapes as Euclidean structures. 
Alternatively, other approaches tried to generalize Convolutional Neural Networks (CNNs) to non-Euclidean manifolds.
Masci \textit{et al. } \cite{masci2015geodesic} introduced Geodesic CNNs that allowed constructing local geodesic polar coordinates that are analogous to patches in images.
Similarly, Boscaini \textit{et al. } \cite{boscaini_learning_2015} proposed localized spectral CNNs to learn class-specific local descriptors based on a generalization of windowed Fourier transform.
This was followed by another generalization of CNNs in \cite{boscaini_learning_2016} denoted as Anisotropic CNNs that replace the conventional convolutions with a projection operator over a set of oriented anisotropic diffusion kernels.
All these approaches allowed extracting local descriptors at each point on deformable shapes, and eventually perform shape correspondence by similarity-matching.

Another category of approaches includes the matching computations in the learning process and can find shape correspondences directly from a CNN.
Litany \textit{et al. } \cite{litany_deep_2017} proposed a structured prediction model in the functional maps space \cite{ovsjanikov_functional_2012} that takes in dense point descriptor for the two shapes, and produces a soft correspondence map.
Halimi \textit{et al. } \cite{halimi_unsupervised_2019} transforms \cite{litany_deep_2017} into an unsupervised setting by replacing the point-wise correspondences with geometric criteria that are optimized, eliminating the need for annotated data.
Donati \textit{et al. } \cite{donati_deep_2020} proposed an end-to-end pipeline that computes local descriptors from the raw 3D shapes, and employs a regularized functional maps to produce dense point-to-point correspondences.
Their method requires less data to train and generalizes better than its supervised counterparts.
Li \textit{et al. } \cite{li2022srfeat} employed a regularized contrastive learning approach to learn robust point-wise descriptors that can be used to match shapes.
We deviate from all these approaches, and we tackle the problem from a peculiar zero-shot perspective that exploits the emerging large-scale models in language and vision.

\subsection{Large-Scale Models}
Several models that are trained on large-scale datasets were introduced recently for different modalities, given the advances in deep architectures design and computational capabilities.
For instance, Large-Scale Language Models (LLMs) such as  T5 \cite{raffel2020exploring}, BLOOM \cite{scao2022bloom}, GPT-3 \cite{GPT3}, and InstructGPT \cite{ouyang2022training}; vision models StableDiffusion \cite{LDM}, and DALLE-2 \cite{DALLE-2}). 
LLMs have outstanding capabilities in understanding textual data, but they lack any understanding of natural images.
Recent methods try to build cross-modal vision-language models that incorporate the capacities of both models.
Visual-ChatGPT \cite{wu2023visual} is one example that combines ChatGPT with many vision foundation models that are managed using a prompt manager that allows better combination and interaction.
MiniGPT-4 \cite{zhu2023minigpt} pursues a similar endeavor by attempting to align frozen LLM with a visual encoder through a projection layer.
To further improve language coherence, they finetune the model on a well-aligned dataset using a conversational template.
BLIP2 \cite{li2023blip2} bootstraps vision-language models through efficient pre-training from off-the-shelf models.
We employ these models to achieve zero-shot 3D shape classification and to generate shape-specific semantic regions that can then be utilized to perform zero-shot 3D semantic segmentation to find shape correspondences.

\subsection{Zero-shot 3D Semantic Segmentation}
Zero-shot 3D semantic segmentation is an active research topic that attempts to segment volumes or point clouds given some textual labels or descriptors \cite{3DGenZ,ZSPCSeg_GeomPrim, PLA,PartSLIP,decatur20233d, zhang2021pointclip,Zhu2022PointCLIPVA,Naeem20213DCZ,abdelreheem2022scanents}.
On a different note, there exist many approaches that are based on Neural Radiance Fields (NeRFs)~\cite{NeRF, NeuralVolumes}, which try to produce full semantic maps of 3D scenes by exploiting 3D density fields from NeRFs \cite{SemanticNeRF, NeSF, NeRF-SOS, PanopticNeRF, PanopticNeRFUrban, NFFF, PanopticLifting}
These two categories of approaches can be combined to perform zero-shot 3D segmentation of volumetric scenes by incorporating zero-shot 2D segmentation networks (\textit{e}.\textit{g}., \cite{LSeg}) into NeRFs \cite{DFF, CLIP-Fields, ISRF} given some textual labels.
SATR \cite{abdelreheem2023satr} showed that replacing 2D segmentation networks with 2D object detector networks yields marginally better results.
Inspired by this, we propose to combine the object detector DINO \cite{DINO} with Segment-Anything (SAM) \cite{kirillov2023segany} to perform zero-shot 3D shape segmentation.

\section{Method}
In zero-shot 3D shape correspondence, the input is a pair of 3D shapes $(S^1, S^2)$, where each shape $S^i$ is represented using triangular meshes with vertices $V^i \in \mathbb{R}^{|V^i| \times 3}$, and faces $F^i \in \mathbb{R}^{|F^i| \times 3}$. 
The number of vertices/faces in $S^1$ are not necessarily equal to that of $S^2$.
The desired output is a point-to-point correspondence map $C \in \mathbb{R}^{|V^2| \times |V^1|}$ that contains matching scores between vertices of $V^2$ and $V^1$.
Note that no other information is provided about the shape such as the shape class or semantic region names, and it is desired to perform shape correspondence in a zero-shot manner with no training or fine-tuning.
To this end, we propose a new setting to tackle this problem that consists of three modules, as illustrated in Figure \ref{fig:method12}.
First, we perform \emph{zero-shot 3D object classification} on the shapes to obtain an object class per shape using a large-scale visual-language model (\ref{class_type_proposal}).
Afterward, a set of \emph{semantic region names} per shape is generated using an LLM (\ref{shared_label_generation}).
Next, \emph{zero-shot 3D semantic segmentation} is performed given the semantic region names (\ref{shape_descriptor_generation}).
Finally, dense correspondence maps can be calculated using functional maps \cite{Ren2018ContinuousAO} (\ref{fmap_computation}).
We explain these components in more detail in the following sections.


\subsection{Zero-Shot 3D Shape Classification}\label{class_type_proposal}

Initially, we need to identify the classes of the 3D shapes.
Existing zero-shot 3D shape classification approaches \cite{cheraghian2019zero,cheraghian2020transductive} can predict a limited set of unseen classes, but do not generalize when the unseen set is unlimited.
In our case, there is no prior knowledge about the classes of the shapes, and therefore, existing approaches for zero-shot 3D classification can not be employed.
To alleviate this, we propose to employ a language-vision foundation model (\textit{e}.\textit{g}., BLIP2 \cite{li2023blip2}) that exploits the generalization capabilities of Large-Scale Language Models (LLMs) to reason about 2D images.

For each shape in the pair $(S^1, S^2)$, we render $k$ views, where viewpoints are sampled uniformly around the shape for a wide coverage.
We set the elevation angles to $\{\ang{-45}, \ang{0}, \ang{45}\}$, the azimuth angles to $\{\ang{0}, \ang{90}, \ang{180}, \ang{270}\}$, and the radius to 2 length units, where each shape is centered around the origin and scaled to be inside a unit sphere.
Then, we feed these $k$ rendered views per shape to BLIP2 \cite{li2023blip2} to obtain $k$ object class proposals.
A natural choice would be to perform majority voting on these predictions to get a single class type. 
However, it is not straightforward to achieve this for textual labels, given that the list of class proposals can include synonyms and adjectives. 
Figure \ref{fig:method12} shows an example of this situation.
Therefore, we exploit the reasoning capabilities of a ChatGPT agent to unify the responses and obtain a single class label per shape.
We show examples of the used prompts in the supplementary materials.


\subsection{Semantic Region Generation and Matching}\label{shared_label_generation}
Zero-shot 3D semantic segmentation approaches require a set of semantic labels as an input together with the 3D shape. 
Our problem is more difficult than traditional co-segmentation, because the two input shapes may not share the same region names.
For this reason, We need to obtain two sets $(R^1, R^2)$ of the possible names of semantic regions present in each shape in the input pair $(S^1, S^2)$. 
Afterward, we attempt to match, whenever possible, between the semantic regions defined in $R^1$ and $R^2$, where a single semantic region in $R^1$ can be matched to one or more semantic regions in $R^2$.
For instance, the legs of a dog can be matched to both the arms and the legs of a person.
We exploit the in-context learning \cite{GPT3} capabilities of LLMs
to achieve this. 
In-context learning is the process by which a model understands a certain task and provides an adequate response to the required task. 
LLM models are indeed good in-context learners, allowing them to perform well on a wide range of tasks without explicit fine-tuning. 
The idea is when asking the model to solve a task given a certain input, we include a few (input, expected output) pairs as examples in the input prompt.
We employ ChatGPT for this purpose, and we query two sets of semantic regions $(R^1, R^2)$ for the two shape classes in question, and a mapping between the two sets $M^{12}: R^1 \leftrightarrow R^2$.
Figure \ref{fig:method12} shows an example of such a mapping.
We refer the reader to the supplementary material for further details on formulating the textual prompts for ChatGPT. 


\begin{figure}[!htb]
    \centering
    \includegraphics[width=0.7\linewidth]{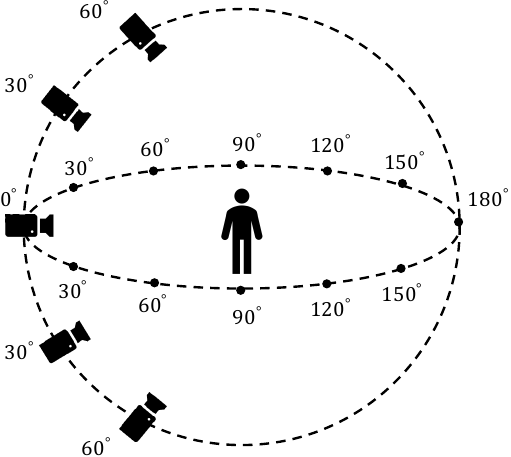}
    \caption[]{Sampling strategy for the $v$ rendered views that are used to perform zero-shot 3D semantic segmentation. Note that we use three different radii of (2, 1.75, 1.5) from the origin to produce a total of 180 views.}
    \Description{This figure shows the camera viewpoints used when rendering an input shape.}
    \label{fig:sphere}
\end{figure}

\subsection{Zero-Shot 3D Semantic Segmentation}\label{shape_descriptor_generation}
After generating the two sets of semantic regions $(R^1,R^2)$, we can use them to perform zero-shot 3D semantic segmentation.
Despite the fact that the recent object segmentation model Segment-Anything (SAM) \cite{kirillov2023segany} is powerful, its text-guided segmentation is still limited. 
The Groudning-DINO object detector \cite{liu2023grounding} on the other hand, can perform 2D object detection for a large number of classes.
Therefore, we propose to combine Groudning-DINO with SAM to perform zero-shot 3D semantic segmentation, and we denote this hybrid approach as \emph{SAM-3D}.
{It is possible to use other object detectors such as GLIP \cite{GLIP} to obtain bounding boxes of parts, but we employ Gounding-DINO as it performs better than its counterparts as shown in \cite{liu2023grounding}.}

We start by rendering a large number of viewpoints $v$ sampled uniformly to cover the whole shape as illustrated in Figure \ref{fig:sphere}. 
For each rendered view, we feed it to DINO to detect a bounding box for each semantic region in $R^i$. 
Afterward, we feed the detected bounding boxes with the rendered viewpoints to SAM to provide segmentation maps for each semantic region.
We define a matrix $X^i \in \mathbb{R}^{|F^i| \times |R^i|}$ that is initialized with zeros, and we use it to accumulate scores for each face in $F^i$ for each semantic region in $R^i$.
Finally, each face is assigned a label by selecting the highest score in each row $j$ of $X^i$ yielding $FL^i$:
\begin{equation}\label{eq:L}
    FL^i = \arg \max_j X^i\left[j,:\right]
\end{equation}

Once the 3D segmentation vectors $FL^i$ are computed, the segmentation maps are matched between the two shapes using the mapping $M^{12}$ to produce a coarse correspondence map.


\subsection{Zero-Shot Dense Shape Correspondence}\label{fmap_computation}
To produce dense point-to-point shape correspondence, {we employ a functional maps-based approach. We use the overall strategy based on associating functional descriptors with region correspondences, as described in the original functional maps paper \cite{ovsjanikov_functional_2012}  and then implemented in \cite{kleiman2019robust} and \cite{ren2018continuous}. 
Specifically, given region-wise correspondences, we formulate an optimization problem to compute a functional map. 
The optimization problem is obtained first by formulating functional constraints using the sum of the WKS descriptor \cite{aubry2011wave} of the points in the segment.
We combine these with the Laplace-Beltrami commutativity regularization into a single system and solve it to find the optimal functional mapping matrix $\mathbf{C}$. 
Finally, we convert the computed functional map $\mathbf{C}$ to a point-to-point map and iteratively refine it using the BCICP refinement strategy  \cite{ren2018continuous}. 
All parameters we use in our approach, including the way we formulate and solve the optimization problem, are exactly the same as in \cite{ren2018continuous} with the only difference being that the regions that are matched are produced by our pipeline rather than the ones produced by \cite{kleiman2019robust}.} 

This gives, as output, a dense point-to-point correspondence between each 3D shape pair.
Interestingly, while functional maps primarily target near-isometric shapes, our initialization with SAM-3D allows us to generate high-quality dense maps even for challenging non-isometric shape pairs, as will be shown in Figure \ref{fig:dense}. 
Nevertheless, artifacts in point-to-point maps can occur due to the use of the functional map framework, and we leave the development of a correspondence densification technique that can adapt to strongly non-isometric shapes, to future work.

\section{Experiments}
In this section, we evaluate our proposed approach, and we introduce our new dataset for \emph{strongly non-isometric}  shape matching (SNIS). 
Moreover, since we propose a new strategy for solving shape correspondence problems, we introduce some evaluation metrics for different components of the pipeline.


\subsection{Strongly Non-Isometric Shapes Dataset (SNIS)}
\begin{figure}[!tb]
    \centering
    \includegraphics[width=\linewidth]{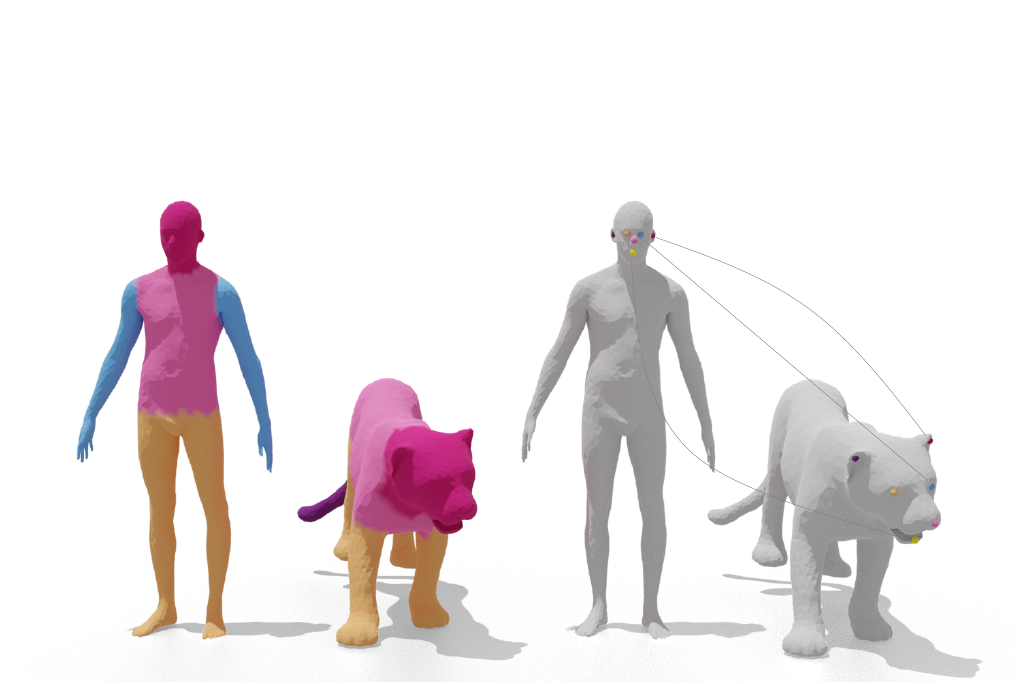}
    \caption[]{Keypoint correspondences and segmentation maps for the proposed Strongly Non-Isometric Shapes (SNIS) dataset. We provide 34 keypoint annotations, but we only show a few here for clarity.}
    \Description{This figure shows an example of the ground truth semantic segmentation for FAUST and SMAL datasets. It also shows 3 keypoints mapping between FAUST and SMAL (the left eye, the left ear, and the mouth).}
    \label{fig:keypoints}
\end{figure}

Existing shape correspondence datasets \cite{Zuffi:CVPR:2017,FAUST} usually encompass a single category of objects (\textit{e}.\textit{g}., humans or animals){, and they employ template models to derive dense correspondences to alleviate annotation workload.}
To facilitate the development of approaches that can generalize to non-isometric shape matching, we introduce a new dataset with mixed shape pairs from existing isometric datasets, \textit{e}.\textit{g}., FAUST \cite{FAUST} (humans), SMAL \cite{Zuffi:CVPR:2017} (animals), and DeformingThings4D \cite{d4dt} (humanoid objects).
For each pair of shapes, we annotate 34 keypoint correspondences between the shapes as well as a dense segmentation map.
Figure \ref{fig:keypoints} shows an illustration for these annotations.
For the FAUST dataset  \cite{FAUST}, {we use a similar approach of annotation as described in} \cite{abdelreheem2023satr} that includes segmentation maps for all the available 100 shapes.

Our SNIS dataset includes 250 shape pairs, where the first shape is either from FAUST or DeformingThings4D, and the second is from SMAL.
The included classes are: \{``cougar'', ``cow'', ``dog'', ``fox'', ``hippo'', ``horse'', ``lion'', ``person'', ``wolf''\}.
Note that {it is desirable to include other categories of objects from diverse datasets such as  SHREC09 \cite{Godil2009SHREC2}. 
Unfortunately, we find that this demands significant manual annotation effort, particularly with point-to-point dense annotation, which is time-consuming and labor-intensive.}
However, we demonstrate the generalization capabilities of our approach by showing some qualitative examples from SHREC09 in section \ref{sec:qual}.


\subsection{Metrics} \label{sec:metrics}
For the final dense shape correspondence map, we use the standard average geodesic error as in \cite{litany_deep_2017, halimi_unsupervised_2019}.
We describe the newly proposed metrics below.

\noindent \textbf{Zero-Shot Object Classification Accuracy (ZSClassAcc)} To evaluate if the predicted object class in \ref{class_type_proposal} is accurate, we compare it against the ground-truth shape label.
However, since LLM-based approaches can predict several synonyms for each class (\textit{e}.\textit{g}., human and person), the standard classification accuracy becomes infeasible.

Therefore, we propose to generate a set of synonyms for each object class in the dataset from WordHoard \footnote{\url{https://wordhoard.readthedocs.io}}.
Whenever a class prediction matches any of the synonyms, it is counted as a correct prediction.
Eventually, the accuracy is calculated as a standard binary classification accuracy.

\noindent \textbf{Semantic Regions Generation F1-Score (SRGen-F1)}
Similar to the previous metric, we evaluate the generated semantic regions as a multi-class classification problem.
Regions that are matched with the ground truth count as True Positives (TP), ground truth regions that were not predicted count as False Negative (FN), and predicted regions that do not exist in the ground truth count as False Positives (FP).
Finally, a standard F1-Score is calculated as:
\begin{equation}
    \text{SRGen-F1} = \dfrac{2 \cdot \text{TP}}{2 \cdot \text{TP} + \text{FP} + \text{FN}}
\end{equation}


\noindent \textbf{Semantic Regions Prediction IoU (SRIoU)}
To evaluate the quality of semantic segmentation for different semantic regions in $R^1$ and $R^2$, we calculate the average intersection-over-union over different regions and shapes as follows:

\begin{equation}
    I^{12} = \dfrac{I^{1} + I^2}{2}
\end{equation}

\begin{equation}
    I^{i} = \frac{1}{|R^{i}|} \ \sum_{r=1}^{|R^{i}|}{IoU^{i}_{r}}
\end{equation}

\noindent where $IoU^{i}_{r}$ is the intersection-over-union for region $r$ in shape $i$ compared to the groundtruth segmentation.


\noindent \textbf{Keypoint Label Matching Accuracy (KPLabelAcc)}
Since we provide keypoint annotations in our proposed SNIS dataset, we can evaluate the shape matching accuracy at these keypoints.
For each shape $i$, we define keypoint indices vector $P^i \in \mathbb{R}^{34\times1}$, which stores vertex indices from $V^i$ for the annotated keypoints.
Given faces labels $FL^i$ from (\ref{eq:L}), we can generate labels for vertices as well that lie on these faces, and we denote them as $VL^i$.
The keypoint label matching accuracy is then calculated between $VL^1, VL^2$, and the groundtruth labels $VL^{GT}$ as:

\begin{equation}
 \text{KPLabelAcc} = \frac{1}{|P|} \ \sum_{j=1}^{|P|}{ VL^1[P^1_j] \ \land \ VL^2[P^2_j] \ \land \ VL^{GT}[P^{GT}_j]}
\end{equation}
where $j$ refers to elements in the vector and $\land$ is the {semantic AND operator, which means the three integers should share the same semantic label.}
This metric measures if the keypoints are matched correctly, and that they were assigned the correct label.
Next, we compare our proposed approach against some baselines in terms of these metrics.


\subsection{Zero-Shot 3D Shape Classification Results}
\textbf{Baseline} We calculate a majority voting between all classification proposals generated by BLIP2.
Table \ref{tab:ZSClassAcc} shows that our proposed approach based on ChatGPT performs significantly better than the standard voting.


\subsection{Semantic Regions Generation and Matching Results}
\textbf{Baseline} We use BLIP2 \cite{li2023blip2} model as a baseline, where we feed it with the $k$ rendered views from section \ref{class_type_proposal} to query semantic regions and mapping.

We report the results in Table \ref{tab:srgenaccTable} for the generated semantic regions in terms of the SRGen-F1 metric.
Surprisingly, our proposed approach that employs ChatGPT outperforms BLIP2 with a huge margin despite the fact that our approach does not have access to the rendered images.
This demonstrates the in-context learning capabilities of ChatGPT.
We can also compare the semantic mapping $M^{12}$ in the same fashion as the semantic regions by matching the keys and values of the generated mapping with those of the ground-truth mapping.
{However, we did not succeed in obtaining a valid mapping from BLIP2 as it accepts only one image at a time, and the prompt is limited to 512 tokens, which can be insufficient for in-context learning prompts.
Therefore, we report only our scores in Table \ref{tab:srgenaccTable}.}


\subsection{3D Semantic Segmentation Results}
We compare against the recently released zero-shot 3D semantic segmentation approach SATR \cite{abdelreheem2023satr} that employs only 2D object detectors.
SATR differs from our approach mainly in using a 2D object detector instead of SAM.

\begin{figure*}[t!]
    \centering
    \includegraphics[width=\linewidth]{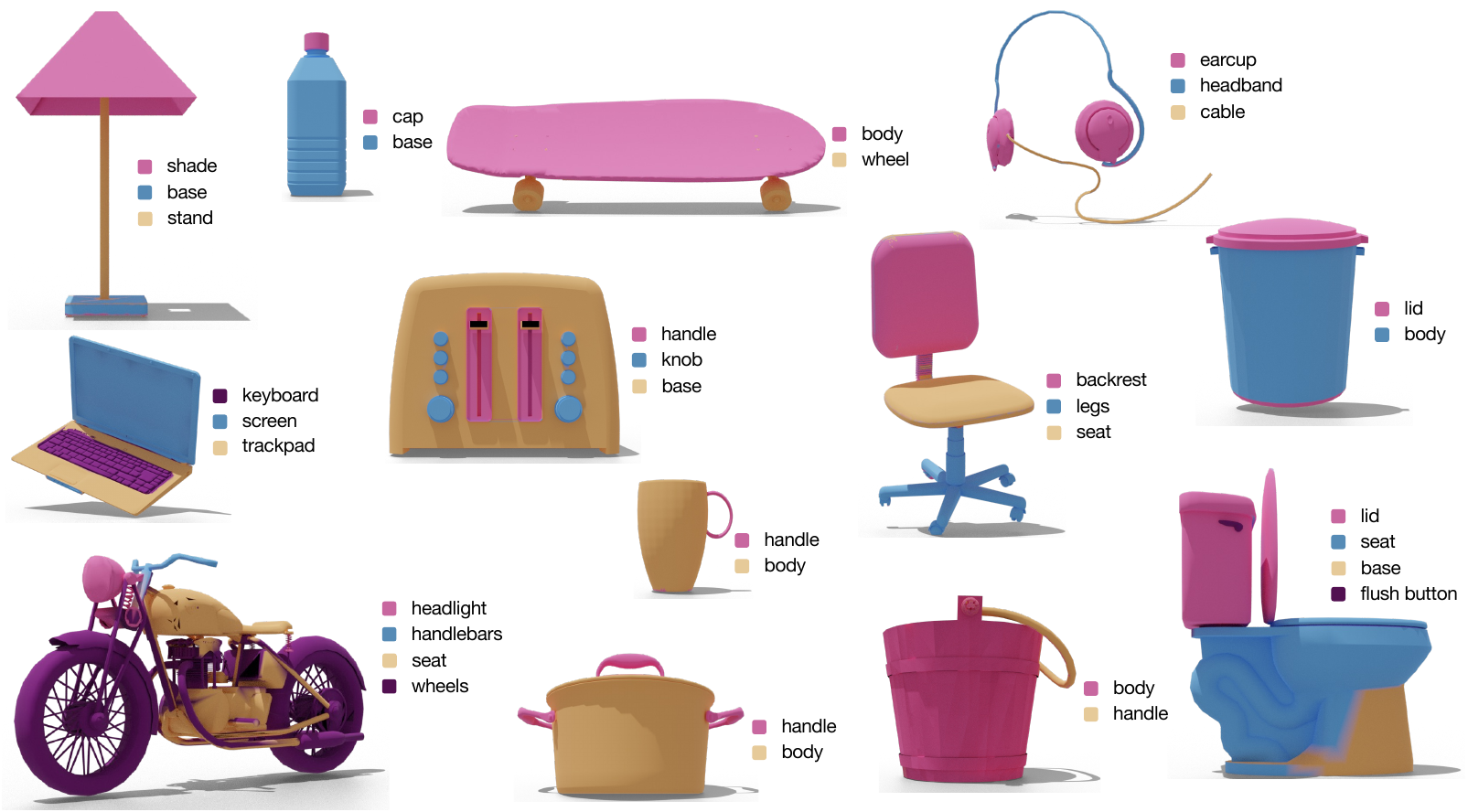}
    \caption[]{{Qualitative examples of SAM-3D for zero-shot semantic segmentation of daily objects. The input textual prompts are provided by ChatGPT. SAM-3D can predict fine-grained parts such as the knobs of a toaster, the cap of a bottle, the flush button of a toilet, or the cable in the headset. }}
    \label{fig:sam3dseg}
\Description{This figure shows the results of SAM-3D for semantic segmentation of daily objects.}
\end{figure*}

Table \ref{tab:seg} shows that our approach outperforms SATR in terms of the SRIoU metric.
We believe that is caused by employing 2D semantic segmentation masks from SAM, which are less prone to error when transferring the segmentation information to the 3D space, compared to bounding boxes.
We also show selected qualitative examples in Figure \ref{fig:satrsam}, where it is clear that our proposed SAM-3D provides a more accurate and well-localized segmentation compared to SATR. {We show in Figure \ref{fig:sam3dseg} the generalization capability of SAM-3D on daily objects.}

\subsection{Keypoints Matching Results}
We compare the keypoints matching results from our proposed approach to those obtained by replacing SAM-3D with SATR \cite{abdelreheem2023satr}.
Table \ref{tab:seg} shows that our approach outperforms SATR in terms of KPLabelAcc, which demonstrates that it provides better segmentation maps with more accurate labels.
\begin{table}[!tb]
    \centering
    \begin{tabular}{lc}
    \toprule
    Method & Acc. \\
    \toprule
    Voting  & 44.80\%  \\
    ZSM{} (ours) & \textbf{73.90\%} \\
    \bottomrule
    \end{tabular}
    \caption[]{{A comparison in terms of Zero-Shot Object Classification Accuracy (ZSClassAcc) between our approach and the standard majority voting.}}
    \label{tab:ZSClassAcc}
\end{table}


\subsection{Dense Shape Correspondence Comparison} 
\begin{figure}[!tb]
    \centering
    \includegraphics[width=1.\linewidth]{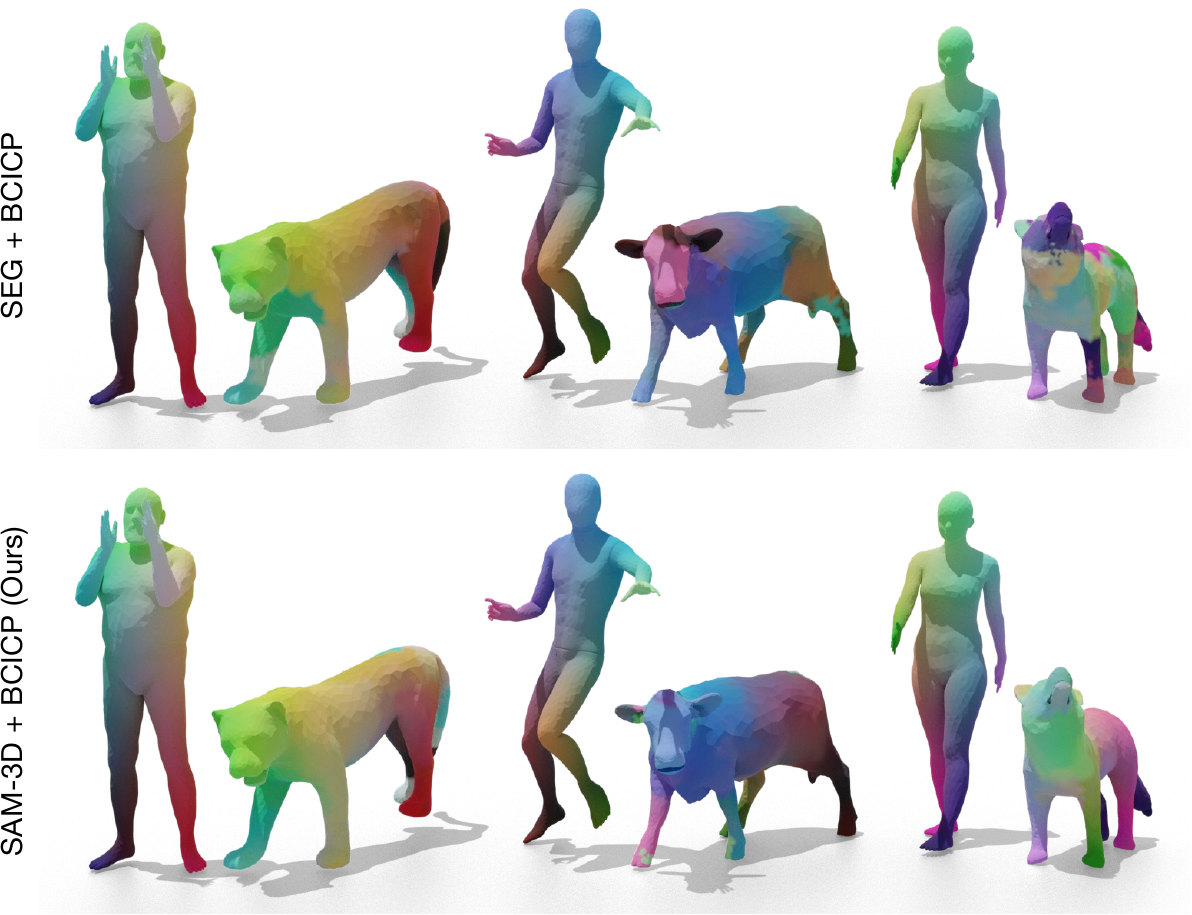}
    \caption[]{Dense shape correspondences generated by the functional maps framework BCICP \cite{ren2018continuous} when initialized with our proposed SAM-3D in comparison to SEG \cite{kleiman2019robust}.}
    \Description{This figure shows qualitative results for dense correspondence between human and animal shapes.}
    \label{fig:dense}
\end{figure}

\begin{figure}[!t]
    \includegraphics[width=1.\linewidth]{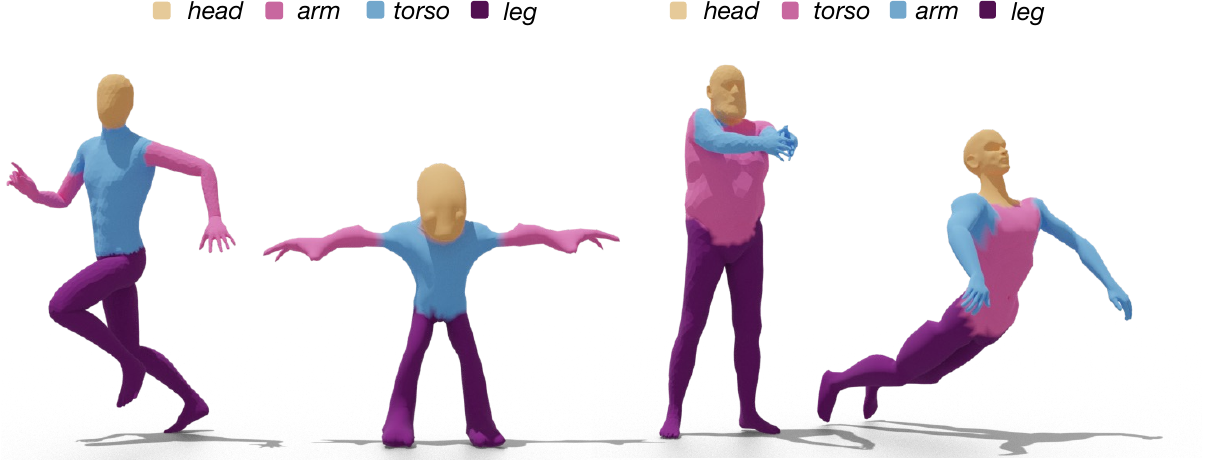}
    \caption[]{Qualitative results for two input pairs of shapes within the same class (left and right columns). The shapes are from DeformingThings4D, FAUST, and SHREC09 datasets.}
    \Description{This figure shows qualitative results for coarse semantic region matching shapes within the same class (humans).}
    \label{fig:figureB}
\end{figure}

Our approach generally produces sparse shape correspondences, as illustrated in {Figures \ref{fig:teaser} and \ref{fig:figureB}}.
However, dense correspondence maps can be produced by using the functional maps framework as described in Section \ref{fmap_computation}.
We provide a comparison for dense shape correspondence maps when using our proposed SAM-3D in comparison with the segmentation model SEG \cite{kleiman2019robust} to initialize the functional maps framework BCICP  \cite{ren2018continuous}
Table \ref{tab:faustCorrespondences} shows that the use of SAM-3D outperforms SEG in terms of average geodesic error with a large margin.
We also provide a qualitative comparison in Figure \ref{fig:dense}, which shows that our approach provides more accurate correspondences, and does not suffer from region discontinuities as SEG. {Note that existing supervised approaches are typically trained on objects of the same category, e.g., humans or animals, where there is enough annotated training data. Therefore, they cannot be directly evaluated on SNIS without adapting and retraining these approaches, which might not always be feasible.}

\subsection{Generalization to Other Datasets} \label{sec:qual}
\begin{figure*}[!htb]
   \begin{minipage}{0.48\textwidth}
    \centering
    \includegraphics[width=\linewidth]{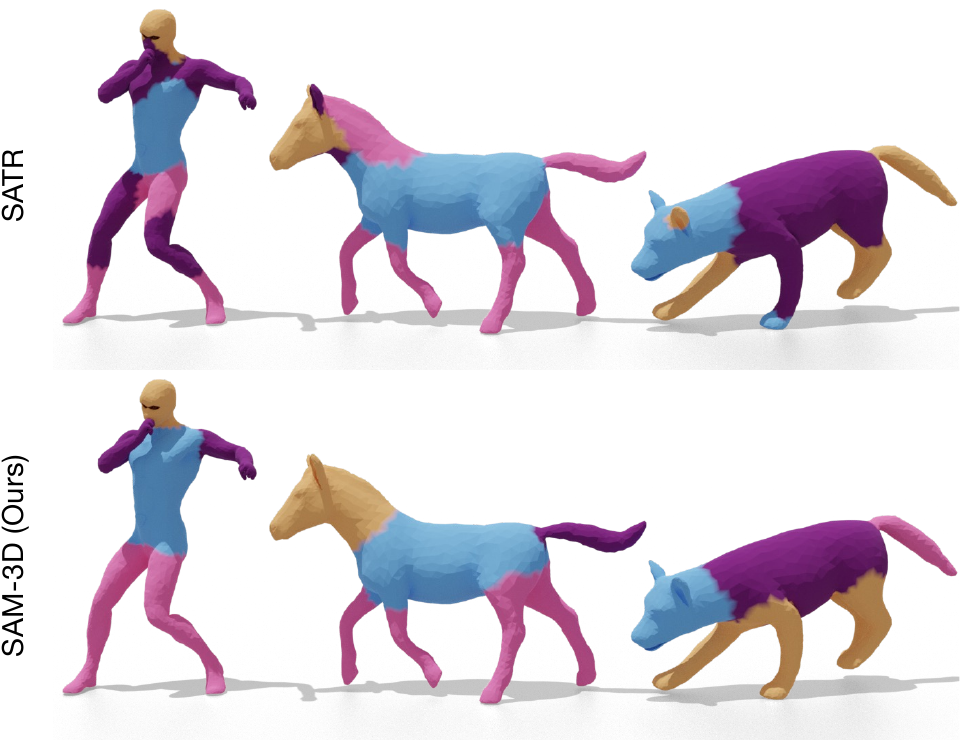}
    \caption[]{Qualitative comparison between our proposed SAM-3D in comparison with SATR \cite{abdelreheem2023satr}. SAM-3D provides more accurate and consistent segmentation compared to SATR.}
    \label{fig:satrsam}
         \Description{This figure qualitative comparison between our proposed SAM-3D in comparison with SATR. Shown examples of a human, a horse, and a dog.}
   \end{minipage}\hfill
   \begin{minipage}{0.48\textwidth}
     \centering
    \includegraphics[width=\linewidth]{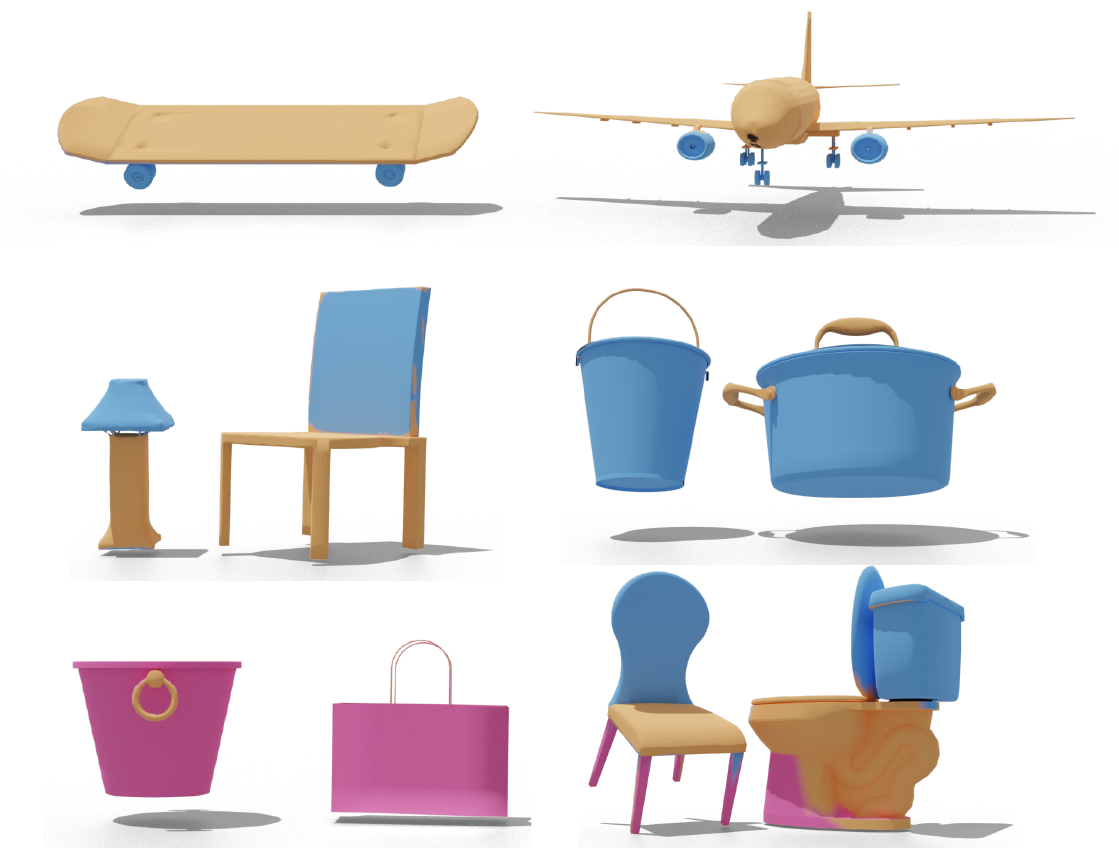}
    \caption[]{{Qualitative examples when matching unrelated daily objects. Our approach produces plausible correspondences demonstrating its reasoning and generalization capabilities.}}
     \Description{This figure shows qualitative examples of our proposed method. A skateboard is matched with an airplane. A lamp is matched with a chair. A kitchen pot matched with a bucket. A bucket matched with a bag. A chair matched with a toilet.}
    \label{fig:generalization2}
   \end{minipage}
\end{figure*}

\begin{figure*}[!t]
    \centering
    \includegraphics[width=0.85\linewidth]{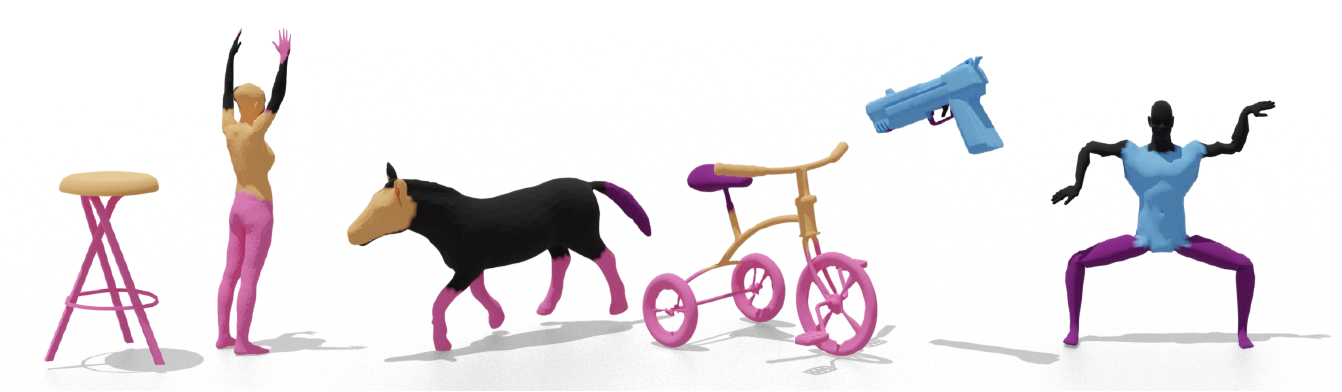}
    \caption[]{Qualitative examples when matching unrelated shapes. Our approach produces plausible correspondences demonstrating its reasoning and generalization capabilities.}
    \label{fig:generalization}
    \Description{This figure shows more qualitative examples of our proposed method.}
\end{figure*}

\begin{figure*}[]
    \includegraphics[width=0.8\linewidth]{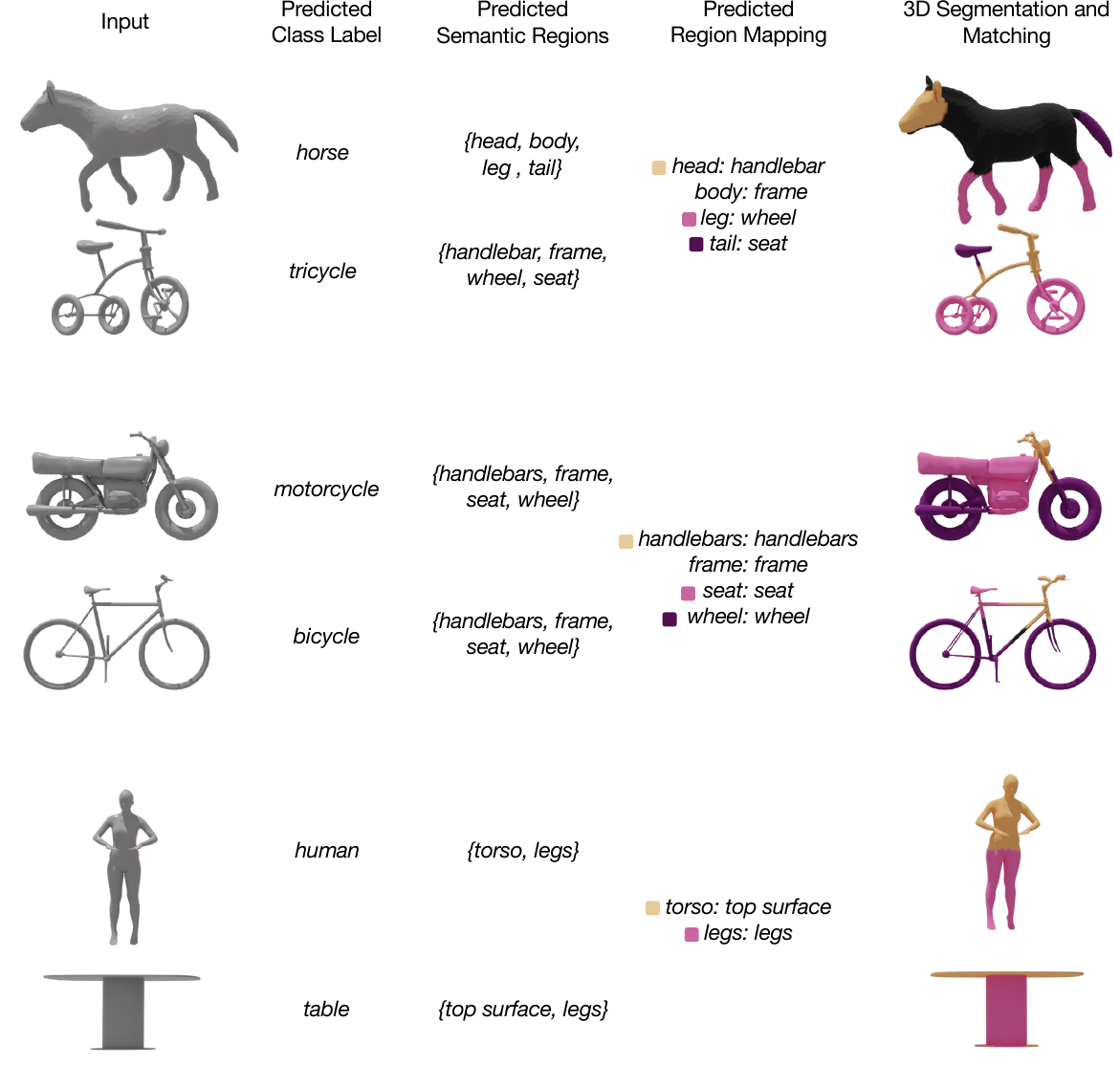}
    \caption[]{Detailed qualitative results for strongly non-isometric pairs of shapes. We show the intermediate predictions from different components of our proposed approach, including the predicted class labels, proposed semantic regions and mapping, and the coarse shape-matching output.}
    \label{fig:detailed}
     \Description{This figure shows detailed qualitative examples of our proposed method.}
\end{figure*}

To examine if our proposed approach generalizes to other datasets with highly unrelated shapes, we include some objects from the SHREC09 \cite{Godil2009SHREC2}, and 3D-CoMPaT \cite{Li20223DCC} datasets.
We form pairs of shapes where the first item is from SNIS, and the second is from SHREC09 or 3D-CoMPaT.
Figures \ref{fig:generalization} and \ref{fig:detailed} show these examples.
Our method was able to produce plausible results when matching a human with a chair, where the legs were matched correctly, and the seat was matched to the rest of the human body.
A horse was also matched to a tricycle, where the horse limbs were matched to the wheels, the head to the handle, and the tail to the seat. {We show in Figure \ref{fig:generalization2} examples where the pairs are from daily objects \cite{ShapeNet,Xiang_2020_SAPIEN}. These examples demonstrate the reasoning capabilities of our approach, even when the shape pairs are not related.}
We provide more detailed qualitative examples in Figure~\ref{fig:detailed}.


\subsection{Impact of Varying Number of Viewpoints}
We examine the effect of changing the number of rendered views $k$, and $v$ in the proposed zero-shot 3D object classifier and in SAM-3D, respectively.
Table \ref{tab:abl} shows that both the classification and segmentation accuracy improve when increasing the number of views.
We do not consider higher values for computational efficiency.


\begin{table}[!tb]
    \centering
    \begin{tabular}{lcc}
    \toprule
    & Semantic Regions & Mapping\\
    Method & Avg. F1-Score & Avg. F1-Score \\
    \toprule
    BLIP2 \cite{li2023blip2} & 41.28\% & -  \\
    ZSM{} (ours) & \textbf{80.31\%}  & \textbf{65.05\%}\\
    \bottomrule
    \end{tabular}
    \caption[]{{A comparison of generated semantic regions and mapping in terms of SRGen-F1. Our proposed approach based on ChatGPT outperforms BLIP2 on semantic region generation. However, we were not able to produce valid mappings from BLIP2, so we only mention ours. }}
    \label{tab:srgenaccTable}
\end{table}

\begin{table}[!tb]
    \centering
    \begin{tabular}{lcc}
    \toprule
    Method & SRIoU & KPLabelAcc \\
    \toprule
    SATR \cite{abdelreheem2023satr} & 69.98\% & 56.60\%  \\
    SAM-3D (ours) & \textbf{73.55\%} & \textbf{59.72\%} \\ 
    \bottomrule
    \end{tabular}
    \caption[]{{A comparison with an existing zero-shot 3D semantic segmentation approach in terms of SRIoU and KPLabelAcc}}
    \label{tab:seg}
\end{table}

\begin{table}[!tb]
    \centering
    \begin{tabular}{lc}
    \toprule
    \ $k$ & ZSClassAcc \\
    \toprule
    1  & 60.10\%  \\
    5  & 62.76\%  \\
    24  & \textbf{73.90\%}  \\
    \bottomrule
    \end{tabular}
    \qquad
    \begin{tabular}{lcc}
    \toprule
    \ $v$ & SRIoU  & KPLabelAcc\\
    \toprule
    
    30  & 69.84\% & 54.97\%  \\
    60  & 72.06\% & 57.63\%  \\
    120  & {73.06\%} & {58.89\%}  \\
    180 & \textbf{73.55\%} &\textbf{ 59.72\%} \\ 
    \bottomrule
    \end{tabular}
    \caption[]{{Ablation study on the effect of changing the number of viewpoints on the zero-shot object classification in terms of ZSClassAcc, and the zero-shot 3D semantic segmentation in terms of SRIoU and KPLabelAcc. }}
    \label{tab:abl}
\end{table}

\begin{table}[!t]
\centering
\begin{tabular}{lc}
\toprule

Method & Avg geodesic error  \\
\toprule
SEG + BCICP \cite{ren2018continuous}  & 0.41  \\
{SATR + BCICP (ours)} & {0.37}\\
SAM-3D + BCICP (ours) & \textbf{0.36} \\
\bottomrule
\end{tabular}
\caption[]{A comparison for dense shape correspondence in terms of average geodesic error. Initializing the BCICP framework with segmentation from SAM-3D yields significantly better results.}
\label{tab:faustCorrespondences}
\end{table}

\section{Conclusion}
We proposed a novel zero-shot approach for 3D shape correspondence.
Our approach exploited the capabilities of recently emerged language and vision foundation models to match challenging non-isometric shape pairs.
There are two key differences in our work to traditional co-segmentation. 
First, we do not require the region names to be known in advance. 
Second, our approach does not require a mutual set of semantic regions and generates shape-specific sets, and a semantic mapping between them instead, enabling it to match diverse shape pairs.
We also introduced a new dataset for strongly non-isometric shapes (SNIS) as well as evaluation metrics for each stage in our pipeline to facilitate the development and evaluation of future methods.

\noindent \textbf{Limitations and Future Work}
Our approach can match coarse semantic regions such as main body parts (\textit{e}.\textit{g}., head, torso, and legs). 
In future work, it would be desirable to produce finer regions in such as eyes, mouth, and hands. 
This is challenging because the current image-based segmentation models are not able to provide fine-grained segmentation for renderings of meshes without textures.
Foundation models in machine learning are helpful for a wide range of tasks. 
In the future, it would also be interesting to design foundation models that can map 3D shapes, images, and text to a common latent space.
Finally, adapting functional maps to handle strongly non-isometric shape pairs, starting from high-quality segment matches, is another interesting problem for future work.


\bibliographystyle{ACM-Reference-Format}
\bibliography{bibliography}

\clearpage

\clearpage
\appendix

\twocolumn[{%
 \centering
 \huge Supplementary Materials for: Zero-Shot 3D Shape Correspondence 
}]

\renewcommand\thefigure{\arabic{figure}}    
\setcounter{figure}{0}
\setcounter{page}{1}

\bigskip
\section{Implementation Details}
We run all our experiments on a single Nvidia RTX 3090 (24 GB RAM). We use the ChatGPT-3.5 turbo model via OpenAI Python API. We use the Nvidia Kaolin library \cite{KaolinLibrary} written in PyTorch for rendering shapes. We render the mesh on a black background with $512\times512$ resolution. We use a bounding box prediction threshold of 3.7 for the DINO \cite{DINO} model.  {To ensure fairness, we use the same number of views when comparing SAM-3D and SATR.}

\begin{figure}[!t]
    \centering
    \vspace{10pt}
    \includegraphics[width=\linewidth]{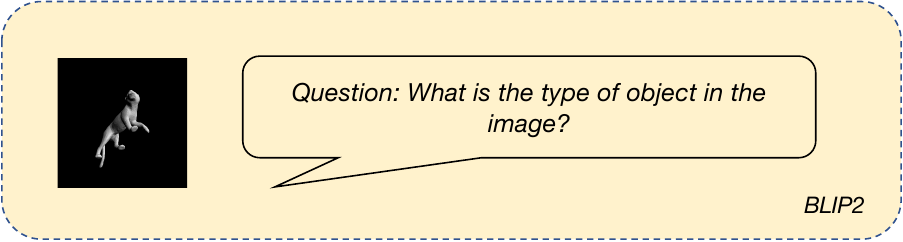}
    \caption{The textual prompt for proposing a class label given a rendered image using BLIP2 model.}
    \label{fig:blipPrompt}
\end{figure}

\section{Semantic Region Generation and Matching Prompts}
\begin{figure*}[!tb]
    \centering
    \includegraphics[width=0.8\linewidth]{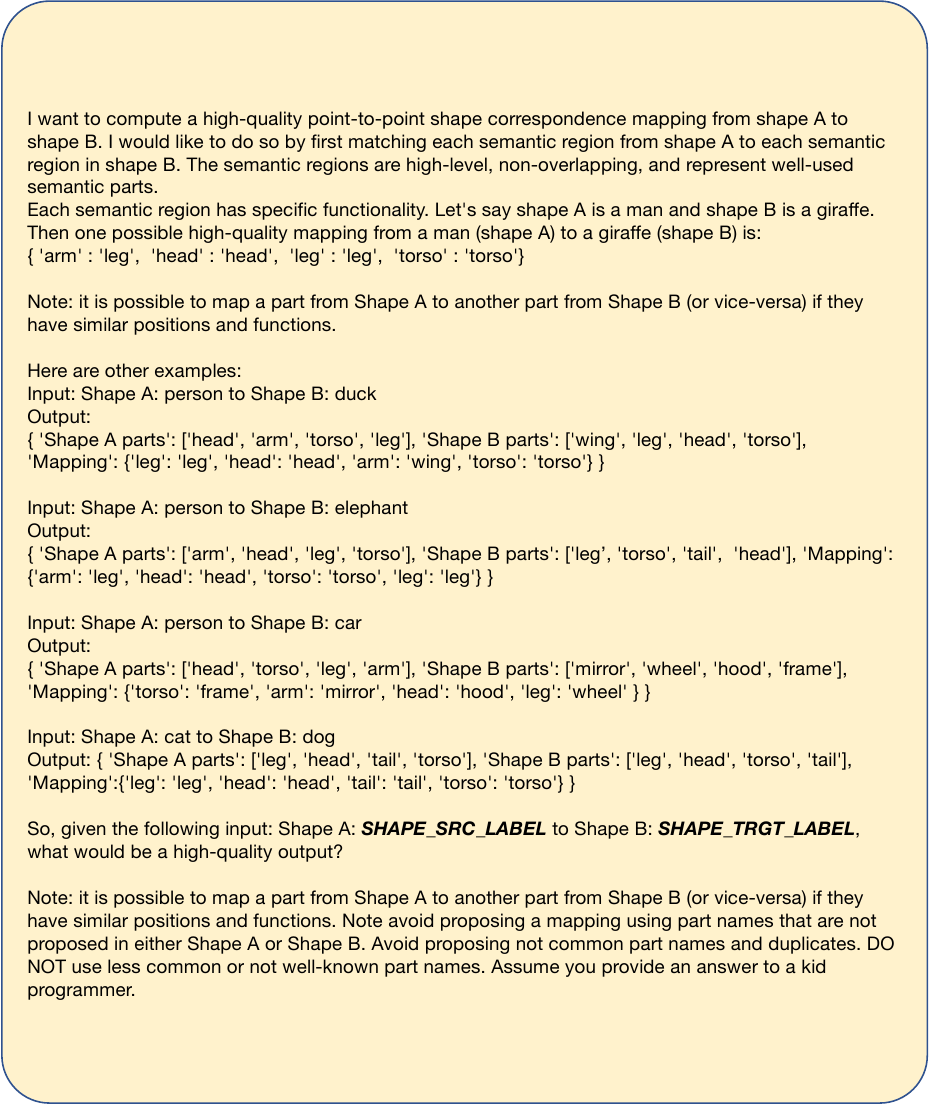}
    \caption{The textual prompt for proposing labels representing the semantic regions for an input pair of shapes and semantic region mapping using ChatGPT agent.}
    \label{fig:chatgptSemanticRegionProposal}
\end{figure*}

In Figure \ref{fig:chatgptSemanticRegionProposal}, we show the textual prompt we use for proposing sets of semantic regions $R^1$, $R^2$ for the input shapes $S^1$ and $S^2$ as discussed in Section \ref{shared_label_generation}. We replace the "SHAPE\_SRC\_LABEL" and "SHAPE\_TRGT\_LABEL" strings with the predicted class label for $S^1$ and $S^2$, respectively.

\section{Prompt Construction Trials}
\begin{figure}[!htb]
    \centering
    \includegraphics[width=\linewidth]{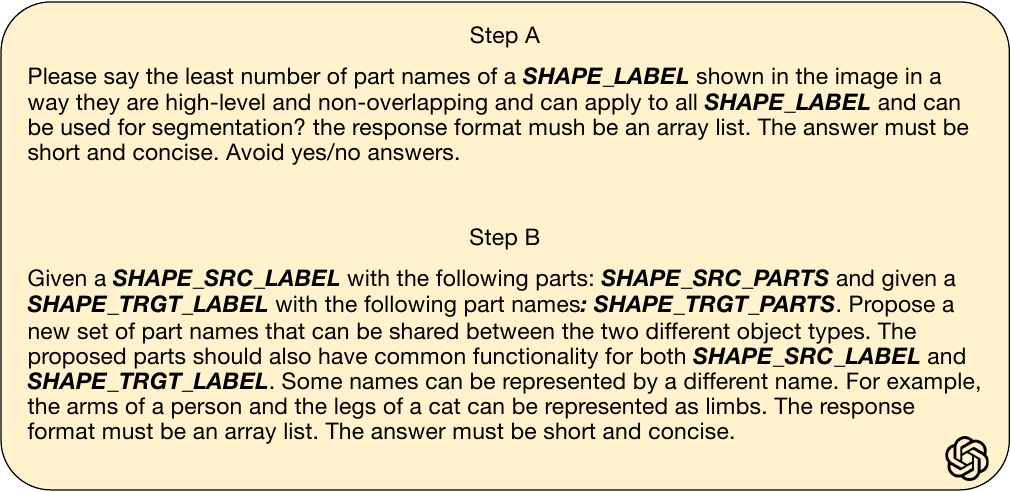}
    \caption{{The two-step textual prompt we used for proposing the semantic region per shape and the semantic region mapping.}}
    \label{fig:albationTextualPrompt}
\end{figure}

{
We investigated different prompts for obtaining the coarse shape correspondences. First, we try a two-step approach. For each shape separately, we ask Visual-ChatGPT \cite{Wu2023VisualCT} to propose a list of semantic regions given at one time in a rendered image. The answers are then unified using ChatGPT in a similar approach as in Figure \ref{fig:chatgptblipPrompt}. Then, we ask ChatGPT to provide a set of semantic regions that can be shared/used for both shapes. We used the prompts, which are shown in Figure \ref{fig:albationTextualPrompt}:}

{So, we construct a better prompt using only a single-step approach as described in Section \ref{shared_label_generation}, wherein the same prompt we ask ChatGPT to provide semantic regions for both input shapes and propose a mapping between the regions. In this manner, we can match region names that are different from each other but can be matched semantically.}

\section{Zero-Shot 3D Object Classification}
\begin{figure}[!tb]
    \centering
    \includegraphics[width=\linewidth]{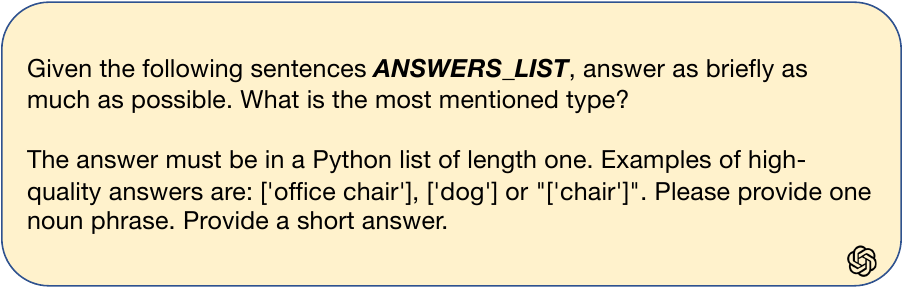}
    \caption{The textual prompt provided to ChatGPT agent to unify the responses produced by BLIP2 model and obtain a single class label per shape.}
    \label{fig:chatgptblipPrompt}
\end{figure}

We show in Figure \ref{fig:blipPrompt} and Figure \ref{fig:chatgptblipPrompt} the prompts used by BLIP2 and ChatGPT in our proposed method for zero-shot 3D object classification. In Figure \ref{fig:chatgptblipPrompt}, we replace the "ANSWERS\_LIST" strings with a list of the proposals predicted by the BLIP2 model given the rendered images of an input shape. 

\subsection{GT Synonyms List}
\begin{figure*}[!tb]
    \includegraphics[width=0.9\linewidth]{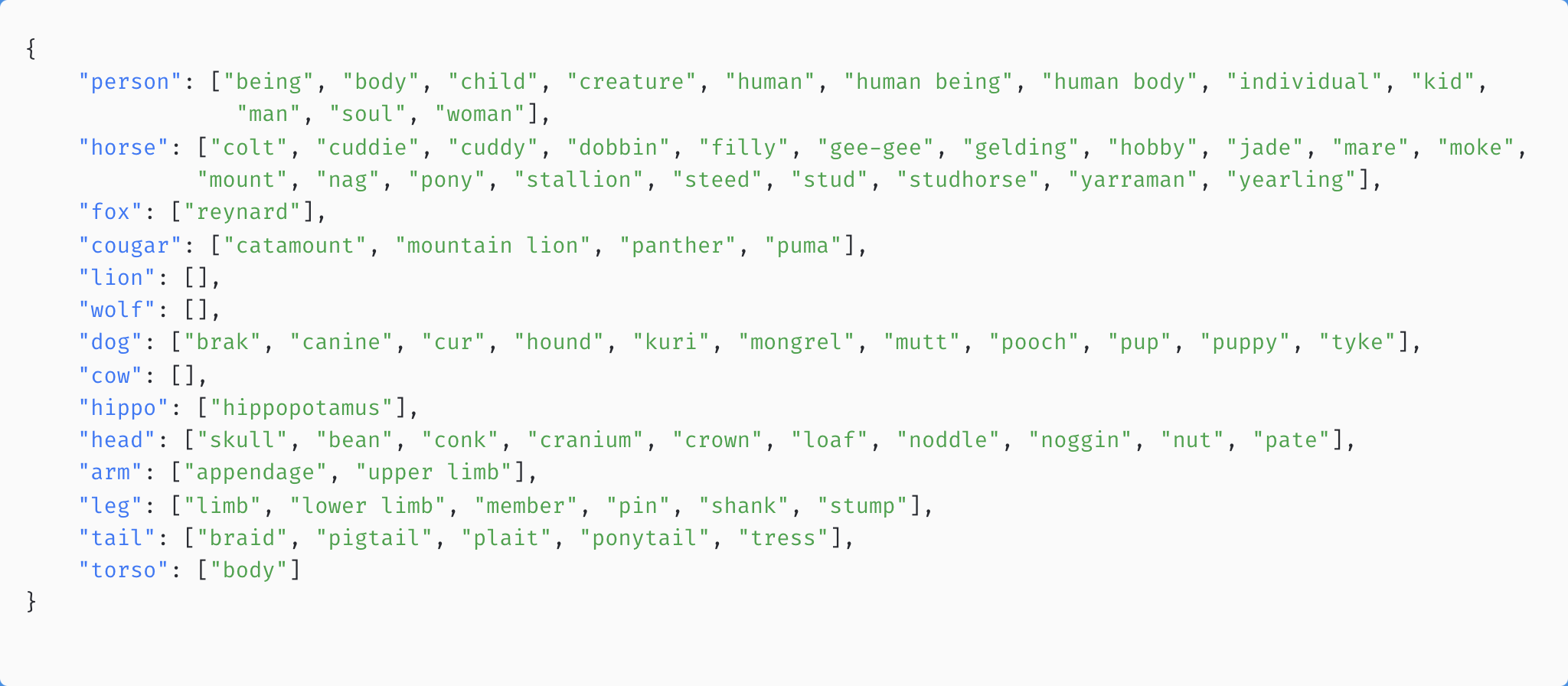}
    \caption{The collected synonyms for the ground-truth object classes and semantic regions we used in our proposed evaluation metrics.}
    \label{fig:syns}
\end{figure*}

Figure \ref{fig:syns} shows the collected synonyms we used in our proposed evaluation metrics.

\clearpage

\end{document}